\documentclass[10pt,twocolumn,letterpaper]{article}

\usepackage{wacv}              

\usepackage{graphicx}
\usepackage{amsmath}
\usepackage{amssymb}
\usepackage{booktabs}
\usepackage{framed,multirow}

\usepackage[pagebackref,breaklinks,colorlinks]{hyperref}

\usepackage[capitalize]{cleveref}
\crefname{section}{Sec.}{Secs.}
\Crefname{section}{Section}{Sections}
\Crefname{table}{Table}{Tables}
\crefname{table}{Tab.}{Tabs.}
\usepackage{xcolor}

\def\wacvPaperID{2135} 
\def\confName{WACV}
\def\confYear{2025}

\begin{document}

\title{
{
%
DMPT: \underline{D}ecoupled \underline{M}odality-aware \underline{P}rompt \underline{T}uning for  Multi-modal Object Re-identification
}
}

\author{
Minghui Lin\textsuperscript{1} \quad 
Shu Wang\textsuperscript{2} \quad  
Xiang Wang\textsuperscript{1} \quad 
Jianhua Tang\textsuperscript{1} \quad  
Longbin Fu\textsuperscript{1} \quad \\ 
Zhengrong Zuo\textsuperscript{1}\thanks{Corresponding author.} \quad
Nong Sang\textsuperscript{1} \\
{\normalsize 
\textsuperscript{1}Huazhong University of Science and Technology}\quad
{\normalsize 
\textsuperscript{2}Shandong University}\\
{\tt\footnotesize \{minghui\_lin, wxiang, techo, fulongbin, zhrzuo, nsang\}@hust.edu.cn, wshu@mail.sdu.edu.cn}
}
\maketitle

\begin{abstract}

{
Current multi-modal object re-identification approaches based on large-scale pre-trained backbones (i.e., ViT) have displayed remarkable progress and achieved excellent performance.
However, these methods usually adopt the standard full fine-tuning paradigm, which requires the optimization of considerable backbone parameters, causing extensive computational and storage requirements.
In this work, we propose an efficient prompt-tuning framework tailored for multi-modal object re-identification, dubbed \textbf{DMPT}, which freezes the main backbone and only optimizes several newly added decoupled modality-aware parameters. Specifically, we explicitly decouple the visual prompts into modality-specific prompts which leverage prior modality knowledge from a powerful text encoder and modality-independent semantic prompts which extract semantic information from multi-modal inputs, such as visible, near-infrared, and thermal-infrared.
Built upon the extracted features, we further design a Prompt Inverse Bind (PromptIBind) strategy that employs bind prompts as a medium to connect the semantic prompt tokens of different modalities and facilitates the exchange of complementary multi-modal information, boosting final re-identification results.
Experimental results on multiple common benchmarks demonstrate that our \textbf{DMPT} can achieve competitive results to existing state-of-the-art methods while requiring only 6.5\% fine-tuning of the backbone parameters.
}

\end{abstract}

\section{Introduction}
\label{sec:intro}
Multi-modal object re-identification (ReID) aims to accurately retrieve objects based on complementary multi-modal inputs, $\eg$, visible (RGB), near-infrared (NIR), and thermal-infrared (TIR). With the development of large-scale benchmarks\cite{zheng2021robust,li2020multi,zheng2022multi,zhu2017unpaired} and foundation models such as ResNet\cite{he2016deep}, Vision Transformer (ViT)\cite{dosovitskiy2020image} and CLIP\cite{radford2021learning}, this field has witnessed unprecedented progress by performing customized adaptations based on these powerful pre-trained backbones.
\begin{figure}[t]
  \centering
   \includegraphics[width=0.98\linewidth]{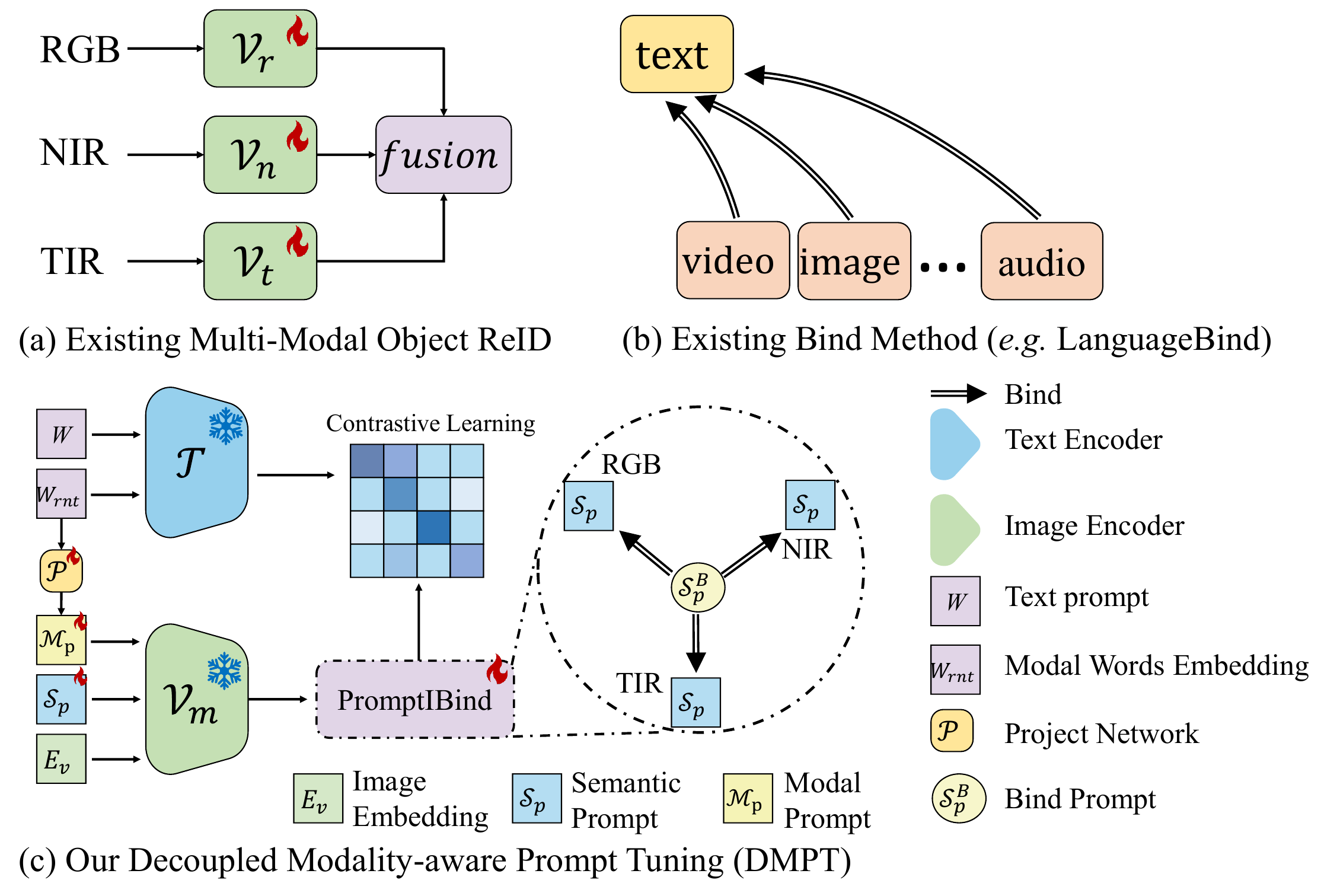}
\vspace{-2mm}
   \caption{(a, c) Comparison of DMPT with traditional multi-modal object re-identification methods. Existing methods use a full fine-tuning architecture for direct feature fusion. In contrast, DMPT employs decoupled interaction based on prompt engineering. (b, c) Comparison of existing bind methods with our PromptIBind-based interaction method. For example, LanguageBind\cite{zhu2023languagebind} uses language as a media for multi-modal alignment. In contrast, we introduce bind prompts for cross-modal inverse interaction.}
   \label{fig:start}
   \vspace{-5mm}
\end{figure}

To explore the complementary cues among different modalities, current methods mainly focus on feature alignment and interaction. For instance,
HAMNet\cite{li2020multi} adaptively combines different spectrum-specific features to achieve heterogeneous fusion and matching of multi-spectral images. PFNet\cite{zheng2021robust} further emphasizes local detail information within global features. IEEE\cite{wang2022interact} designs a multi-modal margin loss to force the model to learn more modality-specific features by enlarging the intra-class feature distance. CCNet\cite{zheng2022multi} proposes a cross-consistency loss to address aligning heterogeneous modalities. GPFNet\cite{guo2022generative} employs a graph structure network to associate features from different modalities, achieving progressive cross-modal fusion. 
Recently, TOP-ReID\cite{wang2023topreid}, EDITOR\cite{zhang2024magic}, and HTT\cite{wang2024heterogeneous} have comprehensively explored the potential of transformers in multi-modal object ReID. 
Despite promising results, these methods still have limitations: (1) The large-scale update and inference processes of fully fine-tuned multi-stream architecture result in substantial computational resource consumption. (2) Some methods\cite{wang2023topreid,li2020multi,zheng2021robust,wang2022interact,guo2022generative,lin2024sanet}, as shown in \cref{fig:start}(a), employ various fusion modules for direct interaction among multi-modal features. However, due to significant modal disparities, such direct interaction may alter the internal feature information of each modality, potentially introducing uncertainties in identity recognition within the intra-modality. Thus, exploring ways to interact with complementary information without changing modal-specific information is a new direction worth investigating.

With the recent development of large foundation models~\cite{radford2021learning,dosovitskiy2020image,wang2024hyrsm++,wang2023cross,zuo2023plip,zuo2024cross}, prompt tuning aimed at adapting pre-trained vision-language models (VLMs)\cite{radford2021learning,wang2023cogvlm,wang2024clip} to downstream tasks started to attract considerable attention as a form of Parameter-Efficient Fine-Tuning (PEFT)\cite{zhou2022learning,hu2021lora,jia2022visual,huang2023vop,liu2024m} due to its simplicity and effectiveness and demonstrated great success.
Inspired by this, we try to incorporate the prompt tuning paradigm into the multi-modal object ReID task and develop an efficient Decoupled Modality-aware Prompt Tuning (DMPT) method tailored for multi-modal object Re-ID in \cref{fig:start}(c).
Specifically, the text prompts are first mapped to the visual-language alignment space as descriptive priors to enhance the visual representation, guiding the network in object-centric feature learning. 
Furthermore, we developed a modality-aware decoupling method to map the modality attribute words in the text prompt to the visual modality prompt and set learnable visual semantic prompts for different modalities in multi-modal image encoders. The method explicitly decouples the visual prompts into modality-specific prompts and modality-independent semantic prompts, which are utilized for prompt tuning of the frozen foundation model.
Finally, based on the decoupling of modality and semantics, we proposed a Prompt Inverse Bind (PromptIBind) strategy, which employs bind prompts as the central medium to mutually couple cross-modal semantic prompts in a "one-to-many" manner. The mutual coupling process includes the external interaction and internal interaction of prompts. 


Based on these improvements, our DMPT is expected to be optimized with few tunable parameters and facilitate the interaction of complementary semantic knowledge without altering the modality-specific knowledge, thereby providing a more powerful multi-modal recognition capability.
Experimental results reveal that DMPT can achieve comparable performance to existing state-of-the-art methods with only $6.5\%$ of the learnable parameters relative to the backbone.

In summary, the major contributions are as follows:

1) We design an effective prompt-tuning framework named DMPT for multi-modal object ReID, which freezes the main backbone and only optimizes several newly added decoupled modality-aware parameters. 

2) A prompt inverse bind strategy is further proposed to explore the complementary cues among different modalities.

3) Extensive experimental results verify the effectiveness of the proposed method, and our DMPT
can achieve competitive results to existing state-of-the-art methods while requiring fewer tunable parameters.



\section{Related work}
\label{sec:relat}
\subsection{Multi-modal object re-identification}\label{mor}
Multi-modal object re-identification (ReID)\cite{li2020multi,zheng2021robust,wang2023topreid,wang2024heterogeneous,cui2024profd} aims to alleviate the generalization difficulties of single-modal under complex visual conditions by leveraging complementary information from multiple modalities. With the successive proposals of datasets such as RGBNT201\cite{zheng2021robust}, RGBNT100\cite{li2020multi}, and MSVR310\cite{zheng2022multi}, multi-modal object ReID has provided new solutions for the visual dilemma of single-modal ReID. However, it has also introduced the problem of heterogeneous learning in multi-modal contexts. To address modal heterogeneity, current methods focus on feature alignment and interaction. HAMNet\cite{li2020multi} adaptively combines different spectrum-specific features to achieve heterogeneous fusion and matching of multi-spectral images, providing a robust baseline for subsequent work. PFNet\cite{zheng2021robust} is designed as a progressive fusion network, revealing the potential of local features in ReID. IEEE\cite{wang2022interact} further emphasizes local detail information within global features and designs a multi-modal margin loss to force the model to learn more modality-specific features by enlarging the intra-class feature distance. CCNet\cite{zheng2022multi} proposes a cross-consistency loss to address the differences between heterogeneous modalities and individuals. GPFNet\cite{he2023graph} employs a graph structure network to associate features from different modalities, achieving progressive cross-modal fusion. Unicat\cite{crawford2023unicat} re-examines the potential modality laziness in multi-modal fusion.

Recently, TOP-ReID\cite{wang2023topreid}, EDITOR\cite{zhang2024magic}, and HTT\cite{wang2024heterogeneous} have comprehensively explored the potential of Transformer in multi-modal object ReID, each focusing on different aspects. TOP-ReID\cite{wang2023topreid} promotes robust fusion and alignment of multi-modal features through a token exchange strategy. EDITOR\cite{zhang2024magic} addresses the negative impact of complex background information by innovatively designing spatial and frequency token selection to suppress background noise. HTT\cite{wang2024heterogeneous} explores the potential of test data, proposing Heterogeneous Test-Time Training to enhance model performance. Although the aforementioned methods have achieved good performance, the fully fine-tuned architecture causes a lot of computational costs. Simultaneously, most approaches focus on the direct fusion alignment of modal features while generally neglecting the exploration of feature contamination caused by heterogeneous information fusion. In contrast, our proposed DMPT freezes the backbone model and can achieve competitive results by only fine-tuning a small number of parameters. In addition, DMPT also emphasizes the independence of modality-semantic information, aiming to learn the complementary information of cross-modality features during interaction.
\subsection{Prompt tuning}\label{prompt}
Prompt tuning is a PEFT paradigm that extends the prior knowledge of pre-trained models to downstream tasks without needing training from scratch. Initially, prompt learning was widely applied in the Natural Language Processing (NLP) field\cite{lester2021power,li2021prefix,qin2021learning}, enabling models to understand language semantics better. With the emergence of large-scale vision-language models (VLMs)\cite{radford2021learning,touvron2023llama,wang2023cogvlm,huang2024learning,liu2025pite}, aligning text with images has become a new paradigm for pre-training. At this stage, the text serves as a prompt to stimulate the model's unified understanding of vision and language. Recently, prompts are no longer limited to text. Visual Prompt Tuning (VPT)\cite{jia2022visual} first extended prompts to the visual domain by adding additional learnable tokens to the frozen model's visual branch and updating the parameters of visual prompts through the gradient descent. MaPLe\cite{khattak2023maple} further proposed multi-modal prompts to achieve optimal vision-language alignment. PromptSRC\cite{khattak2023self} emphasized the importance of historical prompts during training, integrating visual-text prompts from different rounds to improve the model's generalization. PromptKD\cite{li2024promptkd} suggested introducing prompts into the knowledge distillation framework to supervise the student model in learning knowledge from the teacher model. These methods reveal the effectiveness of prompt tuning in downstream tasks, significantly reducing the training resources for models. However, the fine-tuning work for multi-modal object ReID has not been fully explored. Unlike the aforementioned methods, our proposed DMPT is a prompt tuning method tailored specifically for multi-modal object ReID, with a greater focus on the decoupling and coupling of modality-semantic prompts.

\section{Methodology}
\label{sec:method}

\subsection{Model overview}\label{sec:overv}
\begin{figure*}[ht]
  \centering
   \includegraphics[width=0.9\linewidth]{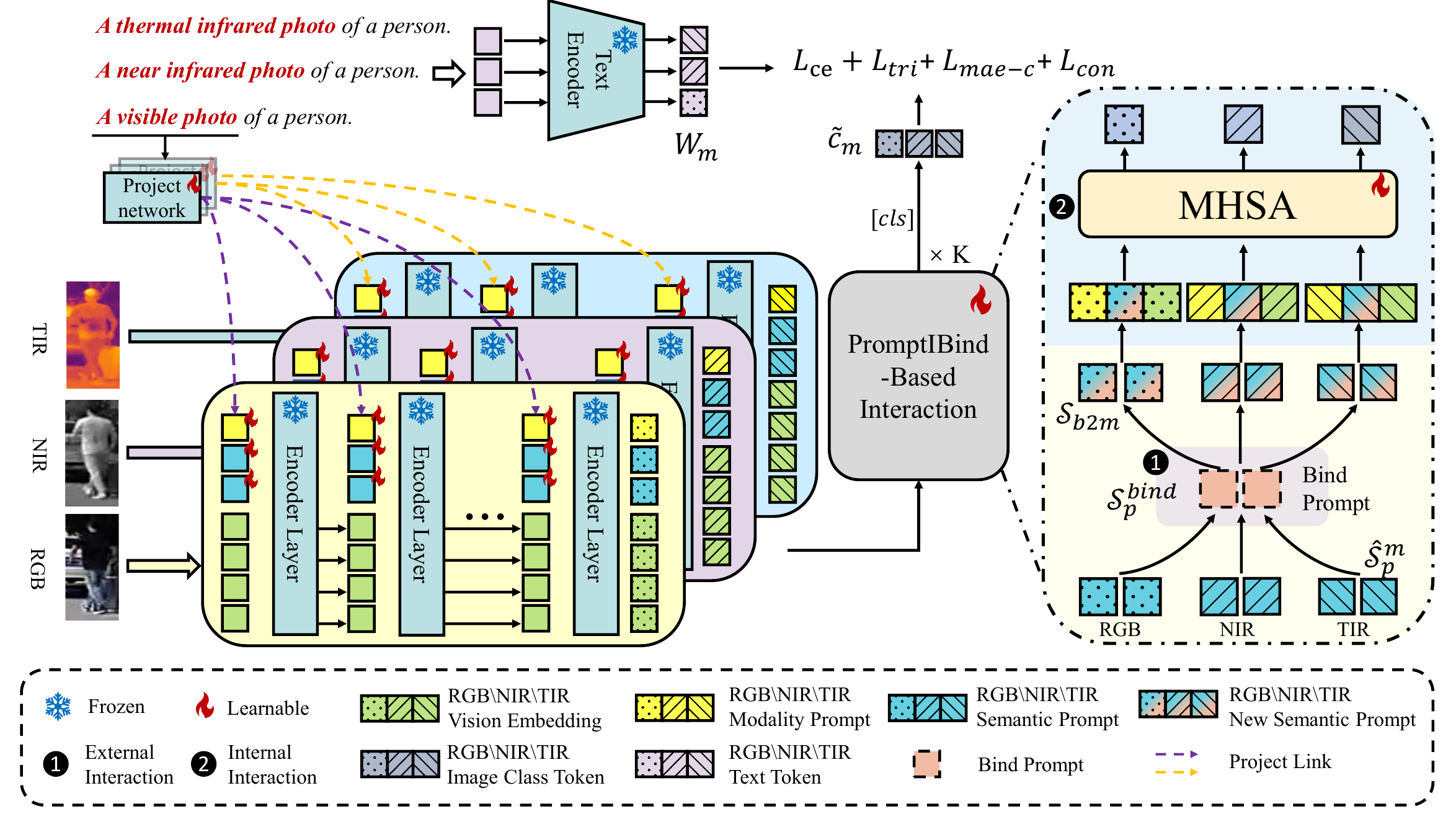}
\vspace{-4mm}
   \caption{Overview of our proposed Decoupled Modality-aware Prompt Tuning (DMPT) framework. We introduce text prompts, modality prompts, semantic prompts, and bind prompts for prompt tuning while keeping the backbone model parameters frozen. Text prompts are aligned with image features to enhance image representation. Visual prompts are decoupled into modality prompts and semantic prompts to further capture cross-modal semantic information. The PrompIBind-based interaction layer achieves cross-modal complementary information interaction synchronously through a "one-to-many" inverse bind structure.
}
   \label{fig:frame}
   \vspace{-2mm}
\end{figure*}

The detailed model process is illustrated in \cref{fig:frame}.  This work employs a frozen foundation model, upon which various specialized prompts are added to construct a prompt tuning architecture DMPT specifically designed for multi-modal object re-identification. Overall, the entire architecture is divided into three parts: modality decoupling, semantic mutual coupling, and auxiliary understanding. In the basic feature extraction layer, the modality attribute words from the text prompts are projected into the visual embedding space to serve as visual modality prompts. At the same time, learnable visual semantic prompts are injected to achieve the decoupling of modality-specific and modality-agnostic information at the prompt level. Subsequently, we obtain basic features from the feature extraction layers for deeper interaction. Here, we propose a prompt inverse bind (PromptIBind) interaction layer, in which we design a bind prompt to connect multi-modal semantic prompts inversely, thereby achieving complementary cross-modal information interaction. Finally, we introduce text prompts aligned with image features to enhance visual representation.
\subsection{Text prompts for multi-modal inputs }\label{sec:textprompt}

The previous CLIP-ReID~\cite{clipreid} method has demonstrated the importance of text concept embeddings for RGB single-modal re-identification tasks. However, in the context of multi-modal object inputs, the recognition of NIR and TIR objects (in addition to RGB objects) relies primarily on structural and contour details. In such cases, using text-based new concept prompts generated by standard gradient optimization clearly lacks semantics and interpretability. For the multi-modal object re-identification task, the key objective is to facilitate the network's comprehension of modality information and semantic information. Therefore, we suggest that the text prompt should not bear the function of advanced semantic comprehension (such as structure, color, or attribute) in this work, but rather should concentrate on object-centric visual-language similarity contrast learning and modality understanding. Thus, we set the text prompts as "a visible/near-infrared/thermal-infrared photo of a person/vehicle."

We adopt the advanced CLIP\cite{radford2021learning} and ViT\cite{dosovitskiy2020image} as the foundation model and follow the setting of VLMs fine-tuning architecture\cite{khattak2023maple}, where the overall framework consists of a text encoder $\mathcal{T}$ and an image encoder $\mathcal{V}$. The inputs of the three modalities—RGB, NIR, and TIR—are configured in a multi-stream architecture, aligned with corresponding text prompts. Initially, the multi-modal images pass through specific embedding blocks to generate the initial patch embeddings ${E_r, E_n, E_t} \in \mathbb{R}^{T_v \times d_v}$, where $T_v$ represents the number of embedding tokens and $d_v$ denotes the dimension of each token. Subsequently, the necessary $[cls]$ token is injected to integrate global information. Similarly, the text prompts are passed through the embedding layer to obtain text embeddings $W_m^l\in \mathbb{R}^{T_t \times d_t}$, $m\in[r,n,t]$ before being input into the text encoder. It is noteworthy that, unlike the above-mentioned image branch, the three text prompts are processed through a shared, frozen CLIP pre-trained text encoder to generate text features. The implementation process is outlined as follows:
\begin{equation}
  [c_r^{l+1},E_r^{l+1}]=\mathcal{V}_r^l[c_r^l,E_r^l]
  \label{eq:1}
\end{equation}
\begin{equation}
  [c_n^{l+1},E_n^{l+1}]=\mathcal{V}_n^l[c_n^l,E_n^l]
  \label{eq:2}
\end{equation}
\begin{equation}
  [c_t^{l+1},E_t^{l+1}]=\mathcal{V}_t^l[c_t^l,E_t^l]
  \label{eq:3}
\end{equation}
\begin{equation}
\left[W_m^{l+1}\right]=\mathcal{T}^l\left[W_m^l\right] 
  \label{eq:4}
\end{equation}
Where, $l\in[0,\mathcal{L}-1]$ represents the hidden layer of each layer of the encoder, $c_{r,n,t}$ denotes the multi-modal $[cls]$ tokens, $[r,n,t]$ represent RGB, NIR, and TIR respectively. 

Then, we obtain vision features and text features of three modalities and map them to the common vision-language (V-L) latent embedding space of each modality through \texttt{ImageProj} and \texttt{TextProj}, represented as $[z_r,t_r],[z_n,t_n],[z_t,t_t]$.

Next, we align V-L features in the shared space, aiming to reduce the text-image gap and enhance object-centric multi-modal representations effectively. The formula for the contrastive loss is as follows:
\begin{equation}
L_{con}^m=\sum_{z_m\in Z}\frac{exp(<z_m,t_m>/\tau)}{\sum_{i=1}^{N}{exp(<z_m,t_m^i>/\tau)}}
  \label{eq:5}
\end{equation}
Where $<\cdot,\cdot>/\tau$ is the cosine similarity calculation with temperature coefficient, $Z$ represents the complete image feature set, and $N$ is the number of object identities. The total hard prompt loss of the three modalities is:
\begin{equation}
L_{con}=L_{con}^r+L_{con}^n+L_{con}^t
  \label{eq:5}
\end{equation}

 \subsection{Modality-aware decoupling}\label{sec:modaldecouple}
One of the essential aspects of multi-modal object re-identification is addressing the dilemma of transferring complementary knowledge across different modalities to enhance the re-identification performance of the model. Previous methods involve direct interaction of original features, which may cause information contamination between modalities, potentially impairing intra-modal identity recognition. We suggest that the content of cross-modal interaction should be discriminative semantic information reflecting the object structure or attributes rather than overall features. 
Therefore, a key insight of our work is to decouple multi-modal features into modality-specific and modality-independent information, allowing for the efficient aggregation of rich semantic information from different modalities without altering the original modality. We observe that in MaPLe\cite{maple}, text prompts and visual prompts undergo tight dynamic knowledge exchange through a coupling function. Inspired by MaPLe, we propose a modality-aware decoupling strategy that utilizes two prompt generation methods to explicitly decouple visual prompts into modality prompts and semantic prompts (modality-independent).

Specifically, we separate the modal attribute words from the text descriptions, expressed as ``a visual/near-infrared/thermal-infrared photo'' and generate modal knowledge tokens through word embedding. Then, these modal tokens are projected into the visual embedding space at each layer of $\mathcal{V}$ via a mapping network $\mathcal{P}$. We call the mapped tokens as the modal prompts $[\mathcal{M}_p^r,\mathcal{M}_p^n,\mathcal{M}_p^t]\in\mathbb{R}^{M \times d_v}$. $M$ is the number of the modal prompt tokens. $\mathcal{M}_p$ implicitly contains the strong modal knowledge from the text description. Meanwhile, we inject randomly initialized learnable visual semantic prompts $[\mathcal{S}_p^r,\mathcal{S}_p^n,\mathcal{S}_p^t] \in \mathbb{R}^{S \times d_v}$ into the multi-modal original visual embeddings, incorporating multi-modal semantic information into the visual prompts. $S$ is the number of the semantic prompt tokens. Explicitly decoupling visual prompts into modal prompts and semantic prompts aids in better harnessing the inherent potential of the semantic prompts in cross-modal interaction. Subsequently, based on the V-L asymmetric framework in \cref{sec:textprompt}, we freeze the visual encoder and inject the learnable visual prompts into the original visual embedding tokens. By fine-tuning with a small number of prompt parameters, the prior knowledge of the backbone is adapted to multi-modal object ReID tasks. The implementation of formula \cref{eq:1,eq:2,eq:3} becomes:
\begin{equation}
[c_r^{l+1},\_,\_,E_r^{l+1}]=\mathcal{V}_r^l[c_r^l,\mathcal{M}_p^{r^l},\mathcal{S}
_p^{r^l},E_r^l]
  \label{eq:6}
\end{equation}
\begin{equation}
[c_n^{l+1},\_,\_,E_n^{l+1}]=\mathcal{V}_n^l[c_n^l,\mathcal{M}_p^{n^l},\mathcal{S}
_p^{n^l},E_n^l]
  \label{eq:7}
\end{equation}
\begin{equation}
[c_t^{l+1},\_,\_,E_t^{l+1}]=\mathcal{V}_t^l[c_t^l,\mathcal{M}_p^{t^l},\mathcal{S}
_p^{t^l},E_t^l]\in\mathbb{R}^{(1+M+S+T)\times d_v}
  \label{eq:8}
\end{equation}

The output of layer $L$-th is:
\begin{equation}
[\hat{c}_m,\mathcal{\hat{M}}_p^m,\mathcal{\hat{S}}_p^m,\hat{E}_m]=\mathcal{V}_m^{L-1}[c_m^{L-1},\mathcal{M}_p^{m^{L-1}},\mathcal{S}
_p^{m^{L-1}},E_m^{L-1}]
  \label{eq:8}
\end{equation}

\subsection{PromptIBind-based interaction}\label{sec:promptibind}

{
In \cref{sec:modaldecouple}, the modality-aware decoupling module introduces two types of learnable visual prompts to divert the additional knowledge learned by the model into modality prompts and semantic prompts. Based on this, the next consideration is how to interact with the high-level semantic prompts of the three modalities effectively.
\begin{figure}[t]
  \centering
   \includegraphics[width=0.88\linewidth]{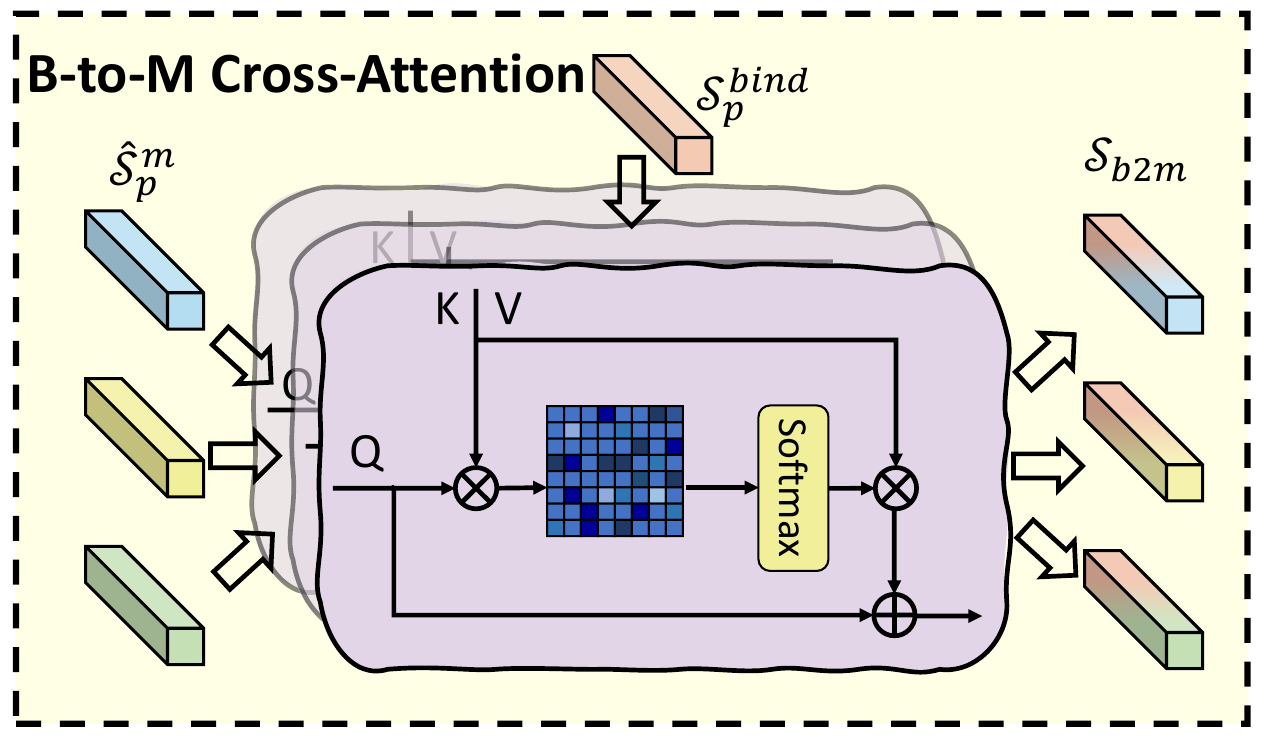}
\vspace{-3mm}
   \caption{Illustration of the bind prompt to modality prompt cross-attention interaction module.}
   \label{fig:b2m}
   \vspace{-5mm}
\end{figure}

Previous approaches have explored coherently aligning elements of diverse modalities by introducing a central modality to bind other different modalities together, such as LanguageBind\cite{zhu2023languagebind}, ImageBind\cite{girdhar2023imagebind}, and OmniBind\cite{wang2024omnibind}, as shown in \cref{fig:start}(b). Since the task of this paper only involves the interaction between prompts, we propose a prompt interaction method named Prompt Inverse Bind (PromptIBind) to connect complementary information across modalities. Intuitively, inspired by the binding structure, we design a prompt located in the pre-alignment space as the center $\mathcal{S}_p^{bind}$ for sharing the semantics of all modalities, defining $\mathcal{S}_p^{bind}$ as the medium of each modality. However, unlike previous bind methods in \cref{fig:start}(b), in \cref{fig:start}(c) we employ a unique inverse bind thinking to interact $\mathcal{S}_p^{bind}$ with each semantic prompt respectively, enabling each modality to receive complementary cross-modal semantic information. Subsequently, the newly acquired semantic prompt information is propagated to the original base features to derive the final multi-modal features. The above interaction layer encompasses external and internal interactions.

{\bf External interaction.} We project the three semantic prompts into a shared space and assign their mean value to $\mathcal{S}_p^{bind}$ without incurring additional training costs. At this stage, $\mathcal{S}_p^{bind}$ completely encapsulates the semantic information from the three modalities, functioning as a medium for their interaction. Subsequently, $\mathcal{S}_p^{bind}$ interacts separately with $[\hat{\mathcal{S}}_p^r,\hat{\mathcal{S}}_p^n, \hat{\mathcal{S}}_p^t]$ through ${\mathcal{CA}}_{b2m}$ (B-to-M Cross-Attention Interaction), with the expectation that each modal semantic prompt acquires complementary information, as shown in \cref{fig:b2m}. Specifically, $\mathcal{S}_p^{bind}$ and $\mathcal{S}_p^m$ are mapped to a shared prompt space through the linear layer, followed by cross-attention interaction between the two. 
Notably, the interaction process among the three modalities and the bind prompt occurs simultaneously and does not involve discrepancies in information from different modalities. The formula implementation process is as follows: 
\begin{equation}
\mathcal{S}_p^{bind}=\frac{1}{N_m}\sum_{m}{W_m^\circ(}\hat{\mathcal{S}}_p^m)
  \label{eq:9}
\end{equation}
\begin{equation}
\mathcal{S}_{b2m}={\mathcal{CA}}_{b2m}\left(\mathcal{S}_p^{bind},\hat{\mathcal{S}}_p^m\right)
  \label{eq:10}
\end{equation}
\begin{equation}
\left({\widetilde{\mathcal{S}}}_p^{bind},{\widetilde{\mathcal{S}}}_p^m\right)=\left(\mathcal{S}_p^{bind}W_b,\hat{\mathcal{S}}_p^mW_m\right)
  \label{eq:11}
\end{equation}
\begin{equation}
\mathcal{S}_{b2m}=\hat{\mathcal{S}}_p^m+\ Softmax(\frac{{\widetilde{\mathcal{S}}}_p^m{({\widetilde{\mathcal{S}}}_p^{bind})}^T}{\sqrt{d_v}}){\widetilde{\mathcal{S}}}_p^{bind}
  \label{eq:11}
\end{equation}
Where $N_m$ represents the number of modals. $W_m^\circ$, $W_m$ and $W_b$ represent different mapping matrices.

{\bf Internal interaction.} To propagate the newly coupled semantic prompts $\mathcal{S}_{b2m}$, which have already obtained cross-modal complementary information, into the basic visual features, we designed the internal interaction layer $\mathcal{SA}$ (Multi-Head Self-Attention)\cite{vaswani2017attention}. We concatenate the original visual features with modality prompts and coupled semantic prompts, then input them into the internal interaction layer, thereby generating enhanced multi-modal features for re-identification tasks. The specific process is as follows:
\begin{equation}
[{\widetilde{c}}_m,{\widetilde{\mathcal{M}}}_p^m,\widetilde{\mathcal{S}}_{b2m},{\widetilde{E}}_m]=\mathcal{SA}_m[\hat{c}_m,\hat{\mathcal{M}}_p^m,\mathcal{S}_{b2m},\hat{E}_m]
  \label{eq:13}
\end{equation}

In the practical process, we design the $k$ layers of the above-mentioned interaction layer $\mathcal{F}_m$ (internal interaction + external interaction) to integrate cross-modal information deeply, summarized as follows:
\begin{equation}
[\widetilde{c}_m^k,\widetilde{\mathcal{M}}_p^{m^k},\widetilde{\mathcal{S}}_{b2m}^k,\widetilde{E}_m^k]=\mathcal{F}_m^k[\hat{c}_m,\mathcal{\hat{M}}_p^m,\mathcal{\hat{S}}_p^m,\hat{E}_m]
  \label{eq:14}
\end{equation}

\subsection{Loss function}
Our method uses cross-entropy loss\cite{szegedy2016rethinking} for labels, triplet loss\cite{hermans2017defense} for features, center MAE loss\cite{wang2024heterogeneous}, and text-image contrastive loss to optimize the model. The extensive cross-entropy loss $L_{ce}$ guarantees that samples of the same identity are correctly classified. The comprehensive application of triplet loss $L_{tri}$ pulls positive samples closer and pushes negative samples further in the feature space. Center MAE loss $L_{mae-c}$ aids in identity recognition within the modality, calculates sample class centers and minimizes the L1 distance between samples and the center. The following is the formula implementation of all objective loss:
\begin{equation}
 L= L_{ce}+L_{tri}+L_{mae-c}+L_{con}
  \label{eq:14}
\end{equation}
\section{Experiments}\label{exper}
\subsection{Datasets and evaluation protocols}\label{data}
    To validate the performance of our method, we evaluated it on four benchmark datasets for multi-modal object re-identification. Two are the multi-modal person datasets RGBNT201\cite{zheng2021robust} and Market1501-MM\cite{wang2022interact}, and the other two are the multi-modal vehicle datasets RGBNT100\cite{li2020multi} and MSVR310\cite{zheng2022multi}. Each image triplet in these datasets consists of RGB, NIR, and TIR. During the evaluation phase, we used mean average precision (mAP) and Rank-1, 5, and 10 reflecting the cumulative matching characteristics (CMC) curve as evaluation metrics.

\subsection{Implementation details}\label{detail}

We applied a multi-modal object ReID task on the pre-trained Transformer from the ImageNet\cite{deng2009imagenet} and the pre-trained ViT-B/16 CLIP model\cite{radford2021learning}. The image sizes for persons and vehicles were adjusted to 256x128. Simultaneously, all images underwent preprocessing techniques such as random horizontal flipping, cropping, and erasing\cite{zhong2020random} for data augmentation. Additionally, modal words were converted into visual modality prompt token $M=1$ by setting projection network $\mathcal{P}$ as a linear layer. The numbers of semantic and bind prompt tokens were both set to $S=32$. During training, we started with a warm-up learning rate of 0.001 and utilized the Adam optimizer with a weight decay of 0.0001 to train the prompt parameters.
\begin{table*}[ht]
  \centering
          \setlength{\tabcolsep}{1.3mm}
{
  \begin{tabular}{@{}llcccccccccc@{}}
    \toprule\toprule
    \multicolumn{2}{c}{{\multirow{2}{*}{Methods}}} &\multicolumn{1}{c}{{\multirow{2}{*}{Backbone}}}&\multicolumn{1}{c}{{\multirow{2}{*}{Params(M)}}}&\multicolumn{4}{c}{RGBNT201}&\multicolumn{4}{c}{Market1501-MM} \\
    \cmidrule(l{0pt}r{0pt}){5-6}
    \cmidrule(l{0pt}r{0pt}){7-12}
    &&&&mAP&Rank-1&Rank-5&Rank-10 &mAP&Rank-1&Rank-5&Rank-10 \\
    \midrule
    \multirow{4}{*}{Single} & HACNN\cite{li2018hacnn} &CNN&-&21.3&19.0&34.1&42.8&42.9&69.1&86.6&92.2\\
     & PCB\cite{sun2018beyond}&CNN&72.3(100\%)&32.8&28.1&37.4&46.9&-&-&-&- \\
     & OSNet\cite{zhou2019omni}&CNN&7.0(100\%)&25.4&22.3&35.1&44.7
&39.7&69.3&86.7&91.3\\
     & CAL\cite{rao2021counterfactual} &CNN&44.5(100\%)&27.6&24.3&36.5&45.7&-&-&-&-\\
    \midrule
    \multirow{8}{*}{Multi} &HAMNet\cite{li2020multi}&CNN&78.0(100\%)&27.6&26.3&41.5&51.7
    &60.0&82.8&92.5&95.0\\
     &PFNet\cite{zheng2022multi}&CNN&-&38.5&38.9&52.0&58.4
    &60.9&83.6&92.8&95.5\\
     &IEEE\cite{wang2022interact}&CNN&109.4(100\%)&46.4&47.1&58.5&64.2 &64.3&83.9&93.0&95.7\\
     &UniCat$\dagger$\cite{crawford2023unicat}&ViT&259.0(100\%)&57.0&55.7&-&- &-&-&-&-\\
     &TOP-ReID$\dagger$\cite{wang2023topreid}&ViT&324.5(100\%)&72.3&76.6&84.7
    &89.4&-&-&-&-\\
     &HTT$\dagger$\cite{wang2024heterogeneous}&ViT&87.7(100\%)&71.1&73.4&83.1 &87.3&67.2&81.5&95.8&97.8\\
     &HTT*\cite{wang2024heterogeneous}&CLIP&87.7(100\%)&75.3&77.3&86.5&90.9&77.4&87.2&97.0&97.6\\
    \cmidrule(l{0pt}r{0pt}){2-12}
    &Baseline$_{1}$&ViT&17.9(6.5\%)&61.8&62.4&75.8&82.9&65.5&83.8&94.4&96.5\\
    &Baseline$_{2}$&ViT&17.9(6.5\%)&62.7&66.0&69.1&85.8&66.8&85.0&94.9&96.9\\
     &{\bf DMPT}$\dagger$&ViT&17.9(6.5\%)&70.9&72.3&81.7&86.3 &69.8&86.6&95.4&97.3\\
    &{\bf DMPT*}&CLIP&17.9(6.5\%)&{\bf 78.5}&{\bf81.3}&{\bf90.4} &{\bf93.5}&{\bf82.7}&{\bf92.0}&{\bf97.1}&{\bf98.5}\\
    \bottomrule
    \bottomrule
  \end{tabular}
  }
  \vspace{-3mm}
  \caption{Performance comparison on two multi-modal person reID benchmarks. The best results are indicated in {\bf bold}. * denotes methods based on CLIP, $\dagger$ denotes methods based on ViT (pre-trained on ImageNet), and the rest are CNN methods. Baseline$_1$ refers to freezing the foundation model and injecting common visual prompts. Baseline$_2$ refers to freezing the foundation model and injecting both common visual prompts and text prompts. (\%) indicates the ratio of learnable parameters to the total model parameters during training.}
  \label{tab:person}
  \vspace{-2mm}
\end{table*}

\begin{table}[ht]
  \centering
            \setlength{\tabcolsep}{3mm}
{
  \begin{tabular}{@{}lcccc@{}}
    \toprule\toprule
    \multicolumn{1}{c}{{\multirow{2}{*}{Methods}}} &\multicolumn{2}{c}{RGBNT100}&\multicolumn{2}{c}{MSVR310} \\
    \cmidrule(l{0pt}r{0pt}){2-5}
    &mAP&Rank-1 &mAP&Rank-1 \\
    \midrule 
    HAMNet\cite{li2020multi}&74.5&93.3&27.1&42.3\\
     PFNet\cite{zheng2021robust}&68.1&94.1&23.5&37.4\\
     CCNet\cite{zheng2023dynamic}&77.2&96.3&36.4&{\bf55.2}\\
     UniCat\cite{crawford2023unicat}&79.4&96.2&-&-\\
     TOP-ReID\cite{wang2023topreid}&81.2&{\bf96.4}&35.9&44.6\\
     HTT\cite{wang2024heterogeneous}&75.7&92.6&-&-\\
    \cmidrule(l{0pt}r{0pt}){1-5}
    {\bf DMPT}$\dagger$&{80.6}&93.2&{35.2} &47.3\\
    {\bf DMPT*}&{\bf 81.7}&94.1&{\bf36.6} &52.1\\
    \bottomrule
    \bottomrule
  \end{tabular}
  }
  \vspace{-2mm}
  \caption{Performance comparison on two multi-modal vehicle reID benchmarks. * denotes methods based on CLIP, $\dagger$ denotes methods based on ViT (pre-trained on ImageNet).}
  \label{tab:vehicle}
  \vspace{-5mm}
\end{table}

\subsection{Comparison with state-of-the-art methods}
{\bf Multi-modal person re-identification.} To verify the effectiveness of the proposed method, we compare the performance of our DMPT with the current state-of-the-art multi-modal object ReID methods on four benchmark datasets. Notably, we use VPT\cite{jia2022visual} with common visual prompts as Baseline$_1$ and additionally introduce text prompts as Baseline$_2$. \cref{tab:person} reports the experimental comparison of various methods on person datasets RGBNT201\cite{zheng2021robust} and Market1501-MM\cite{wang2022interact}. The CLIP-based DMPT achieved notable improvements of 6.2\% and 15.5\% mAP on two datasets, respectively, compared to state-of-the-art TOP-ReID\cite{wang2023topreid} and HTT\cite{wang2024heterogeneous} methods. The ViT-based DMPT also improved by 2.6\% on the Market1501-MM dataset, demonstrating the effectiveness and strong generalization ability of our method. In addition, we compared our method on two backbones with the most promising HTT\cite{wang2024heterogeneous} by replacing the original HTT backbone ViT with CLIP. The results show that the CLIP model under multi-modal pre-training performs better than the ViT pre-trained on ImageNet. Furthermore, 
our DMPT has demonstrated a significant and comprehensive improvement under both backbones, effectively proving the advancement of our method. Compared to the two baselines that rely solely on rudimentary prompts, our DMPT achieved superior performance, highlighting the importance of modality-aware decoupling and PromptIBind-based interaction.

{\bf Multi-modal vehicle re-identification.} As shown in \cref{tab:vehicle}, our method still performs exceptionally well on the RGBNT100 and MSVR310 datasets. On the RGBNT100 dataset, the mAP and Rank-1 accuracy are 81.7\% and 94.1\%, respectively. On the MSVR310 dataset, our method obtained a mAP of 36.6\% and a Rank-1 of 52.1\%. This demonstrates that our method can robustly adapt to different datasets.

{\bf Parameter efficiency.} As shown in \cref{tab:person} and \cref{tab:vehicle}, we achieve superior performance on four benchmark datasets compared to fully fine-tuned methods. We achieved results comparable to full fine-tuning methods (100\% parameters) by utilizing only 6.5\% of the total model parameters on the same backbone, even outperforming the majority of full fine-tuning methods. DMPT saved up to 93.5\% of training memory usage, establishing it as the most memory-efficient model among all existing methods for multi-modal object re-identification. These findings clearly suggest that our approach introduces promising prospects for the efficient implementation of multi-modal ReID.

\subsection{Ablation studies}
{\bf The effectiveness of prompt components.} To investigate the contribution of each prompt component in DMPT, we conducted ablation experiments, as shown in \cref{tab:ablation}. The following experiments all employ CLIP as the backbone model. The four prompt components are visual semantic prompts $\mathcal{S}_p$, visual modality prompts $\mathcal{M}_p$, bind prompts $\mathcal{S}_p^{bind}$, and text prompts $W_t$. When only the semantic prompts $\mathcal{S}_p$ were added, DMPT exhibited remarkable performance, demonstrating the effectiveness of prompt learning in the context of re-identification. Notably, the semantic prompts here function as ordinary visual prompts\cite{jia2022visual} in the absence of other components. We then introduced bind prompts to verify the impact of interactions with ordinary visual prompts on the results. It can be observed that compared to \cref{tab:ablation}(2), the 
\begin{table}[t]
  \centering
              \setlength{\tabcolsep}{3.5mm}
{
  \begin{tabular}{@{}l|llll|cc@{}}
    \toprule
&$\mathcal{S}_p$&$\mathcal{M}_p$&$\mathcal{S}_p^{bind}$&$W_t$&mAP&R-1 \\
    \midrule 
    (1)&$\checkmark$&$\times$&$\times$&$\times$&71.7&74.4\\
     (2)&$\checkmark$&$\times$&$\checkmark$&$\times$&73.5&76.9\\
     (3)&$\checkmark$&$\checkmark$&$\checkmark$&$\times$&76.7&80.6\\
     (4)&$\checkmark$&$\checkmark$&$\checkmark$&$\checkmark$&78.5&81.3\\
    \bottomrule
  \end{tabular}
  }
  \vspace{-2mm}
  \caption{Ablation of semantic prompt $\mathcal{S}_p$, modality prompt $\mathcal{M}_p$, bind prompt $\mathcal{S}_p^{bind}$ and text prompt $W_t$ of DMPT on RGBNT201 dataset.}
  \label{tab:ablation}
  \vspace{-5mm}
\end{table}
introduction of bind prompts achieved a small improvement, emphasizing the necessity and effectiveness of the PromptIBind. Then, we proceeded to introduce visual modality prompts $\mathcal{M}_p$. At this stage, the image information was successfully decoupled into modality and semantic information, performing PromptIBind interactions. It can be observed that the decoupling of information improved performance from 73.5\% mAP to 76.7\%, which fully demonstrates the importance of modality-semantic decoupling before interaction. This observation provides evidence that direct feature interactions may lead to feature contamination. Finally, the text prompts brought additional textual information, improving performance by 1.8\%, making the model more focused on the object region within the images. Thus, the four prompt components complement each other, resulting in a significant performance enhancement for multi-modal object ReID.

{\bf Interaction layer depth.} \cref{tab:interact} shows the impact of the number of interaction layers on performance. Without interaction (0 layers), the model performs standard prompt fine-tuning, showing no significant performance improvement. With one interaction layer, the interaction layer receives the encoder's output token sequence and introduces the bind prompts to effectively facilitate semantic interaction under decoupled conditions, demonstrating the effectiveness of PromptIBind. However, adding more layers weakens decoupling behavior, potentially contaminating cross-modal information and decreasing performance. Therefore, a single interaction layer achieves optimal results.

\begin{table}[t]
  \centering
  \setlength{\tabcolsep}{3mm}
{
  \begin{tabular}{@{}l|cccc@{}}
    \toprule
Interaction Layer Depth&mAP&R-1&R-5&R-10 \\
    \midrule 
    \multicolumn{1}{c|}{{0}}&72.1&75.9&86.1&89.2\\
    \multicolumn{1}{c|}{{1}}&78.5&81.3&90.4&93.5\\
     \multicolumn{1}{c|}{{2}}&75.2&79.2&86.6&90.2\\
    \bottomrule
  \end{tabular}
  }
  \vspace{-3mm}
  \caption{Ablation of the PromptIBind-based interaction layer depth on RGBNT201 dataset.}
  \label{tab:interact}
  \vspace{-4mm}
\end{table}

\begin{table}[t]
  \centering
  \setlength{\tabcolsep}{7mm}

{
  \begin{tabular}{@{}l|cc@{}}
    \toprule
Modality Prompt Length&mAP&R-1\\
    \midrule 
    \multicolumn{1}{c|}{{0}}&75.1&77.2\\
    \multicolumn{1}{c|}{{1}}&78.5&81.3\\
     \multicolumn{1}{c|}{{2}}&77.2&82.1\\
     \multicolumn{1}{c|}{{3}}&75.6&78.9\\
    \bottomrule
  \end{tabular}
  }
  \vspace{-2mm}
  \caption{Ablation of the modality prompt length on RGBNT201 dataset.}
  \label{tab:mp-len}
  \vspace{-5mm}
\end{table}

\begin{figure}[t]
  \centering
   \includegraphics[width=0.8\linewidth]{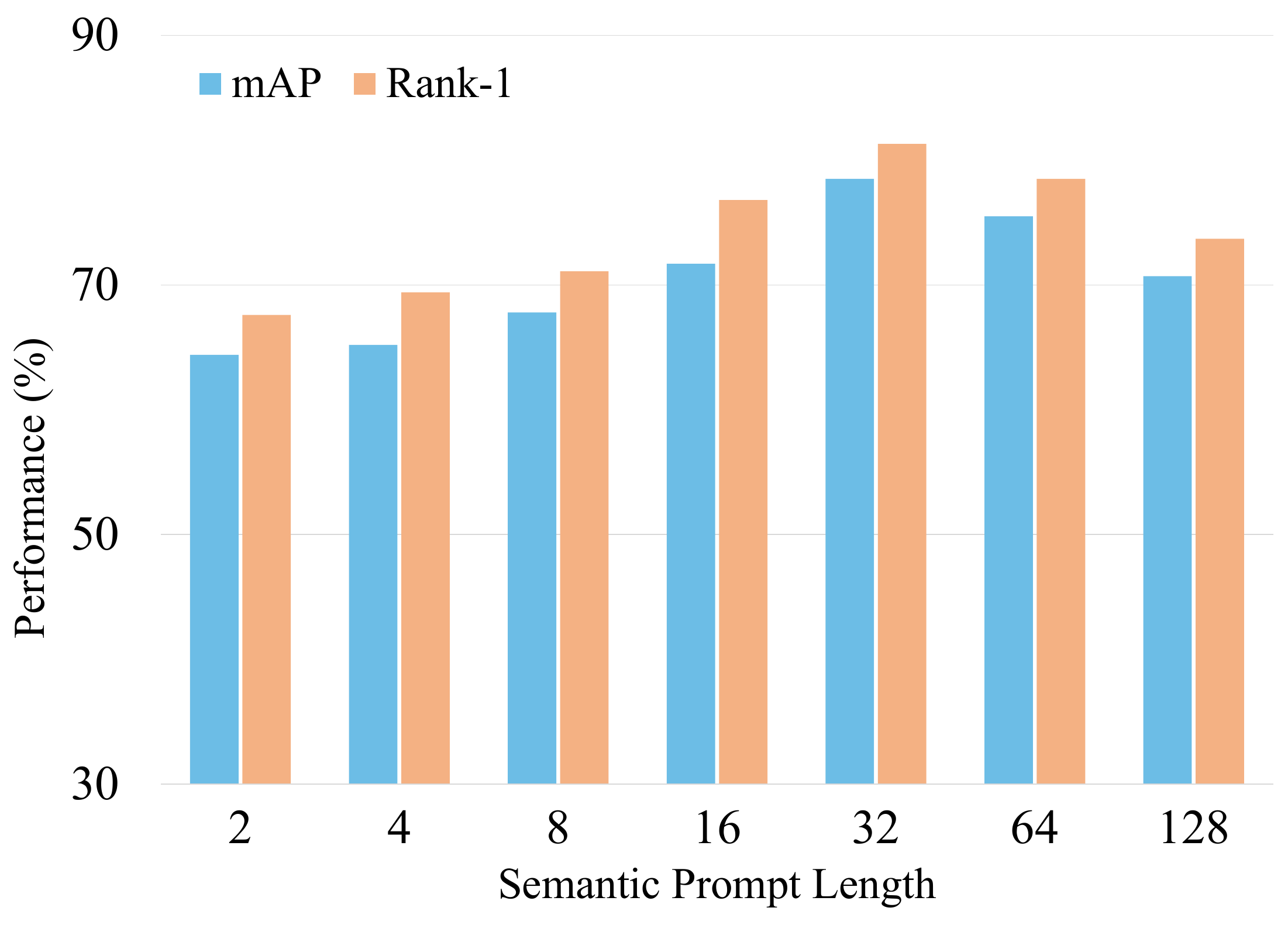}
\vspace{-3mm}
   \caption{Ablation of the semantic prompt length. We select the best semantic prompt length for our DMPT.}
   \label{fig:promptlen}
   \vspace{-5mm}
\end{figure}

{\bf Prompt length.} As a hyperparameter that needs to be learned in addition to the interaction layer during the entire training process, the length of the prompt is crucial to the model's performance. As shown in \cref{fig:promptlen} and \cref{tab:mp-len}, taking the RGBNT201 dataset as an example, we conducted hyperparameter experiments on different lengths of modal prompts and semantic prompts. The optimal semantic prompt length is 32, and the optimal modal prompt length is 1.

\begin{figure}[t]
  \centering
   \includegraphics[width=0.98\linewidth]{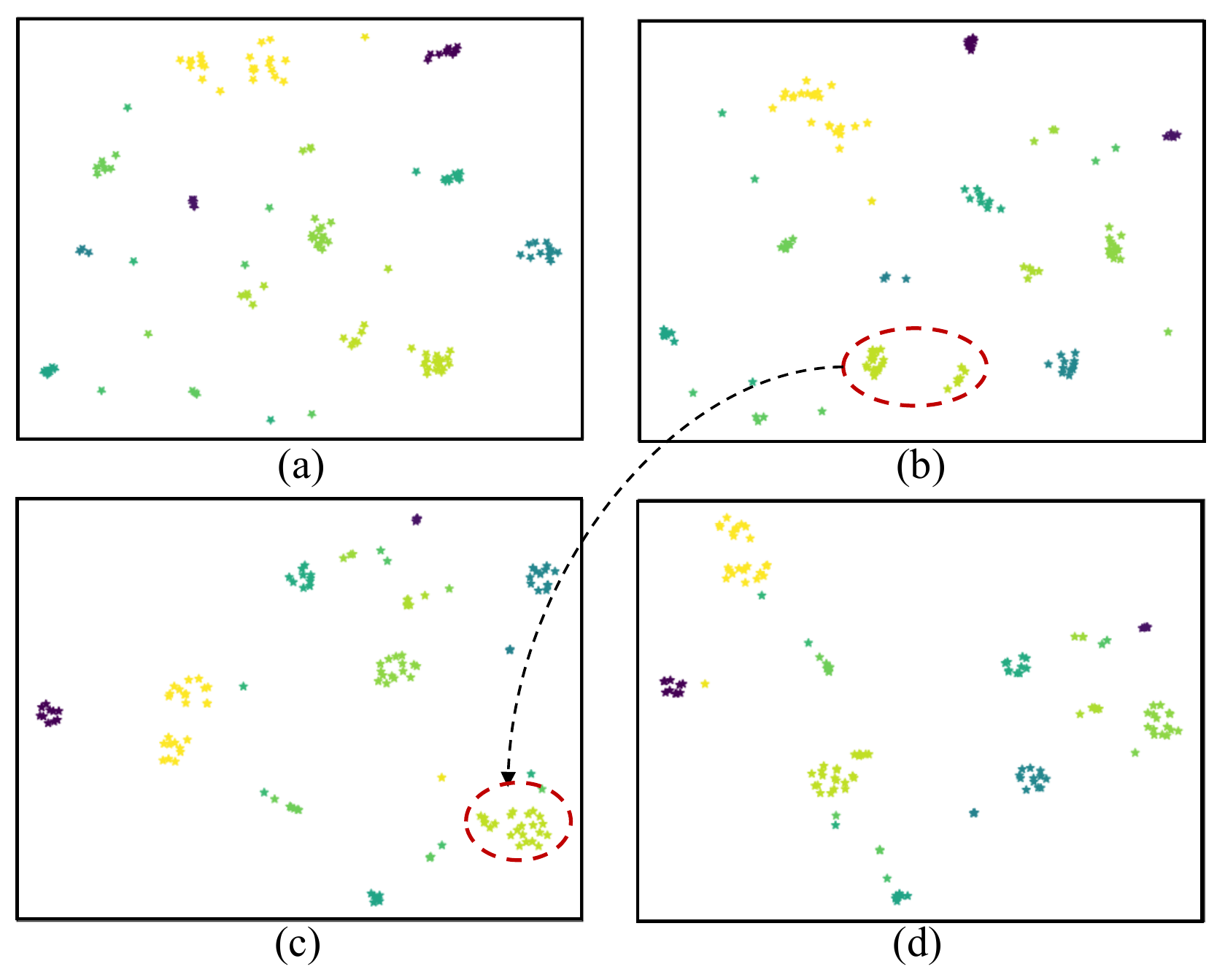}
\vspace{-3mm}
   \caption{t-SNE visualization of features. Different colors represent different identities. (a) $\mathcal{S}_p$(baseline$_1$); (b) $\mathcal{S}_p$+$\mathcal{S}_p^{bind}$; (c) $\mathcal{S}_p$+$\mathcal{M}_p$+$\mathcal{S}_p^{bind}$; (d) $\mathcal{S}_p$+$\mathcal{M}_p$+$\mathcal{S}_p^{bind}$ +$W_t$. }
   \label{fig:tsne}
   \vspace{-2mm}
\end{figure}

\subsection{Visualization}
{\bf Feature distribution.} In \cref{fig:tsne} visualizes the feature distributions obtained by adding different prompt components, corresponding to \cref{tab:ablation}. Compared to the baseline method as shown in \cref{fig:tsne}(a), the introduction of bind prompts for interaction in \cref{fig:tsne}(b) somewhat clusters the previously scattered features, although some remain scattered. We further introduce modality-aware decoupling in \cref{fig:tsne}(c), using only semantic prompts for interaction, which results in a tighter distribution of features with the same ID. Finally, in \cref{fig:tsne}(d), adding text prompts enhances the representative capability of the features, resulting in a clearer feature distribution. By visualizing the feature space distribution, all the prompt components show intuitive improvements compared to the baseline.

\section{Conclusion}
In this work, we propose DMPT, a novel prompt-tuning framework tailored
for multi-modal object re-identification. We introduced a modality-aware decoupling module to effectively decouple visual prompts into modality-specific and modality-independent semantic prompts, and proposed the PromptIBind strategy, which utilizes bind prompts to synchronously exchange cross-modal complementary information. During fine-tuning, we froze the foundation model and updated the prompt parameters. By updating only 17.9M model parameters (6.5\% of full fine-tuning), DMPT achieved competitive performance compared to full fine-tuning, enabling efficient fine-tuning for multi-modal object re-identification.

{\bf Acknowledgments} This work is supported by the National Natural Science Foundation of China under grants 623B2039 and U22B2053.

{\small
\bibliographystyle{ieee_fullname}
\bibliography{egbib}
}

\end{document}